# A Survey on Fundamental Concepts and Practical Challenges of Hyperspectral images

[1] Hasna Nhaila*, [3] Elkebir Sarhrouni, [4] Ahmed Hammouch

Electrical Engineering Research Laboratory, ENSET
Mohammed V University
Rabat, Morocco
[1] hasnaa.nhaila@gmail.com, [3] sarhrouni436@yahoo.fr, [4] hammouch_a@yahoo.com

*Abstract*-The Remote sensing provides a synoptic view of land by detecting the energy reflected from Earth's surface. The Hyperspectral images (HSI) use perfect sensors that extract more than a hundred of images, with more detailed information than using traditional Multispectral data. In this paper, we aim to study this aspect of communication in the case of passive reception. First, a brief overview of acquisition process and treatment of Hyperspectral images is provided. Then, we explain representation spaces and the various analysis methods of these images. Furthermore, the factors influencing this analysis are investigated and some applications, in this area, are presented. Finally, we explain the relationship between Hyperspectral images and Datamining, and we outline the open issues related to this area. So, we consider the case study: HSI AVIRIS 92AV3C. This study serves as map of route for integrating classification methods in the higher dimensionality data.

## I. INTRODUCTION

Remote sensing with hyperspectral images uses the atmospheric transmission of electromagnetic radiation. This transmission is particularly high in the areas of visible (400 nm 700 nm), near infrared (700 nm 1300 nm) and shortwave infrared (1300 nm 3000 nm). Other areas are frequently used in remote sensing, but for most applications of spectroscopic imaging, only these three areas are interesting.
The principle of remote sensing is based on the observation that the Earth's surface and objects react differently to solar radiation (passive sensing), depending on the type of materials and their physical conditions (humidity, etc...). Plants for example have different spectral characteristics in the visible and near infrared. Thus, this science aims to exploit the information contained in the spectral signature to identify objects. From this point of view, hyperspectral remote sensing is an important revolution. It allows to collect numerous and detailed information on the spectral signatures of the objects observed.

A. *Definition of Hyperspectral Imaging*

A generic definition of HSI was made by Kruse [1] as: "Hyperspectral imaging consist of to acquire spectra for all image pixels, where a spectrum is a contiguous measure of the wavelength distribution with sufficient resolution to resolve the natural variability of the system of interest".
Hyperspectral sensors pick up the signals in a very broad spectrum and different parts of the spectrum can have different capacity to distinguish objects of interest: The intrinsic spectral distinction of different objects are not necessarily identical in the same wavelengths or bands. In some parts of the spectrum, the materials may have a much more nuanced spectral reflectance than in others. In addition, the complex transmission conditions in the atmosphere (Bands untransmuted), such as water and absorption of C02 also play a role in this phenomenon.

In the same context of HSI definition, Chang [2] cites three main advantages of HSI regarding the multispectral images MSI:

*1) The number of bands in HSI is more than a hundred.*

while the multispectral contains just three at ten images.

*2) The bands* are regularly spaced in HSI, but those of multispectral images are irregularly spaced.

*3) The spectral resolution:* (central wavelength divided by the width of spectral band) is about a hundred against ten for multispectral images.

B. *Acquisition Process and Treatment of HSI*

Usually, the most widely used as a source of illumination is the sun so we have the passive remote sensing systems. The sensors measure the light reflected (or emitted energy) from areas of interest. The data is finally transmitted to the ground for manual analysis (by expert) or automatically.

*1) Acquisition Process of HSI*

In general, a remote sensing system can be divided into three main components: the scene, the sensors and processing algorithms [3]. Modeling scene includes solar lighting (which covers the area of solar spectral reflectance), the atmospheric transmittance, contiguity effects, shadow effects and clouds.

Other problems can be illustrated by the variation of the scene perspective when it is observed from different angles. The model of the sensors includes radiometric noise sources such as shot noise, thermal noise, readout noise of the detector, quantization noise and calibration error. The processing component comprises atmospheric compensation, various linear transformations, and a number of operators used for the distribution of data.

From technical point of view, two approaches are used for HSI acquisition [4]. The fust is to acquire a sequence of two-dimension images at different wavelengths (staring system), using variable filter positioned in front of a camera matrix.
The second approach is to acquire a sequence of images line by line, such that for each pixel of a line, a complete spectrum is measured (push-broom approach). The spatial dimension is acquired by relative movement of the sensor to the object.

*2) Treatment of HSI*

The HSI preprocessing contains four steps:

*a) The image calibration with respect of the sensor noise:* it varies with the meteorological conditions. Every day, an image called (dark image) is taken to obtain data relating to sensor noise. To minimize the noise of the sensor the raw image undergoes, with this dark image, a processing called minimum noise fraction (MNF) [5].

*b) Correction of geometric distortion:* caused by the movement of the platforms under different atmospheric disturbances.

*c) Geo-registration of the image:* using a triangulation method with bilinear resampling [6]: Fifteen to twenty points of ground control (GCP: ground control points) are distributed inside and outside the bounds and are used to georeferenced each image.

*d) The calibration of the image:* respecting the variation of the illumination using a method called the empirical damping (empirical line method) [7].

## II. CASE STUDY: HSI AVIRIS 92AV3C

NASA uses the "Airborne Visual and Infra-Red Imaging Spectrometer" (AVIRIS). It acquires 220 images called bands in the spectral range from 0.4 to 2.45 microns. Each band has the size of 145x145 pixels. The Ground Truth map used in the experiments are acquired from the AVIRIS sensor (AVIRIS 92AV3C) [8]. Each image is of size 145x145, Two-thirds of the stage are covered by agricultural land and one third by forests, other building structures can also be seen in the scene. Each pixel is labeled as one of the 16 vegetation classes or unidentified. Figure.1,
The availability of reference data makes this excellent hyperspectral image source for the realization of experimental studies [9].

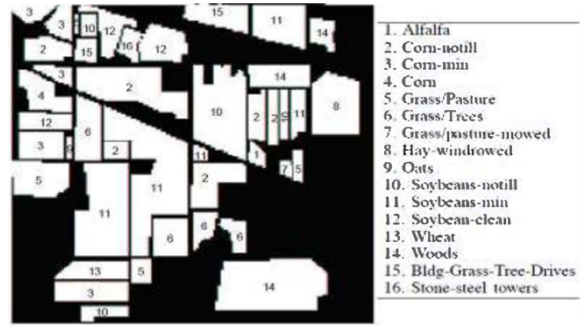

Fig. 1. The Ground Truth map of AVIRIS 92AV3C

At certain frequencies, the spectral images are known to be affected by atmospheric water absorption. Some studies eliminate them [9]. Figure 2 shows the spectral reflectance of the classes '9', '14' et '16', extracted from the HSI AVIRIS 92AV3C. The axis x indicates the number of spectral bands (1-220), and axis y represents the pixel value in the measurement of the different bands. Significant overlap between the two classes occurs in some bands due to natural variability and similarity of spectral reflectance, A class '9 ' for example is " embedded " in the two others. To separate them, we must consider their statistical characteristics, such as Means [Figure 2(b)] and standard deviation [Figure 2(c)] for each spectral band. In other bands, e.g the 60-65 range, these classes overlap broadly.

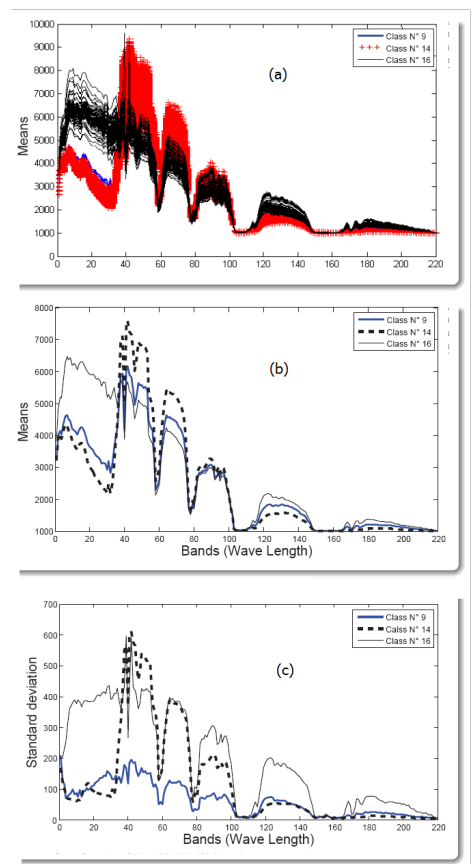

Fig. 2. The non-uniform distribution of the information in the HST. (a)Samples of spectral responses for three classes of vegetation in AVIRIS 92AV3C:"16","14" and"9", the statistical characteristics of the spectral reflectance values in each spectral band, (b) Means, (c) standard deviation.

Figure 2 also illustrates that in hyperspectral imaging, discriminatory information is not distributed evenly across the spectrum. Among all spectral bands, some may contain more useful information for classification than others, and therefore they have great separability indices. Whereas the measure of separability gives an estimated probability of correct classification that benefit from the most informative subsets.

### III. REPRESENTATION SPACES OF HYPERSPECTRAL DATA

One of hyperspectral imagery characteristics is that each pixel is defined by a vector whose elements are the different spectral components (wavelength) from the captured scene. This hyperspectral vector provides not only color information but also information regarding the chemical composition of the materials in the scene. The hyperspectral image can be seen as a data cube where each pixel is a vector of dimension equal to the number of bands. The quasi-continuous nature of the spectrum measurement introduces a strong local correlation of bands. This inter-band correlation expresses the data redundancy.

#### A. Image Space

This is the most intuitive way to represent the data tapes of HSI. This is a geographical representation of substances reflections of the scene, in the frequency band. The weak point of this representation is that the relationship between the bands is not apparent [10], its strong point is that it allows to take into account the spatial dependencies pixels.

#### B. Spectral Space

This is a presentation of measures for a pixel, depending on wavelengths, its basic idea is to represent the distribution of information relating to a point in the field, according to wavelength that transmit information necessary to identify the contents of a pixel, which is interesting from economic view of point. This identifies each pixel independently of their neighbors. This representation gives direct interpretations and in association with the physical properties of the pixel content. In this representation, we can separate a reduced number of classes even with a single band. This type of method does not take into account the relationships between neighboring pixels, so it does not address the texture [8].

#### C. Attributes Space or Features Space

With a given class of substance, this representation is interested in their reflectance in two different bands, and generally in a space of dimension equal to the number of wavelengths. The advantage of this is that it gives a diagnostic on how the reflectance of a pixel are distributed around their average, and so the possibility of using data mining algorithms [10]. Also, the classes whose spectral representation does not distinguish can be separated in the attribute space.

#### D. Generalization

A study on HSI may be done by a combination of the three afore mentioned spaces, in the sense of the use of information that can be extracted in each field. Masalmah [11] indicates that HSI processing algorithms can be grouped into three types: algorithms for spectrum only those of spatial processing, and spaciospectral algorithms. Natahlie [12] presents an iterative combination of spatial and spectral methods in the segmentation of hyperspectral picture.

### IV. ANALYSIS METHODS OF HSI DATA

Much of the research on HSI aims to find more effective ways to make profits for this type of data. Thus, recent research methodologies can be classified into two types, Zhang [13]: The Pure-Pixel methods and Mixed-Pixels methods. This is common to all three above representation spaces of hyperspectral data.

#### A. Pure-Pixels Methods

These methods are based on the assumption that each pixel of any band is composed of a single substance, and consequently it is defined by a unique signature. These methods can be grouped into two categories: Vegetation Index methodology and statistical methods.

*1) Vegetation Index Methods:* The vegetation index is extracted from the pixel spectrum measured by the HSI, then it is compared with the real measurements saved in a library. The main problem of the Vegetation Index is that it is difficult to construct an index of universal vegetation suitable for most hyperspectral data [13] and the variability of spectral signatures for the same type of vegetation.

*2) The Statistical Methods:* In these methods each band is *considered* as a random variable, in which the statistical methods are applied to extract statistics features of the images. These methods require dimensionality reduction to reduce the computational cost, because of the difficulty to apply the statistical models directly on the HSI. One of the most important applications of these methods is the detection of anomalies.

#### B. Mixed-Pixels Methods:

There are two factors omitted in the approaches of Pure-Pixel methods: The complexity of the field (overlapping vegetation etc..) and the limitation of sensor resolution of HSI. These methods are designed to overcome these problems. They are classified in two types: linear mixture models, and nonlinear mixture models.

*1) Linear mixture models:* they present a clear physical meaning the proximity of substances. Zhang [13] indicatesthat the number of classes to be retrieved must be lower than the number of bands in the HSI.

2) *Nonlinear mixture models:* they are used to deal with the problem of the limited number of relevant bands; The mixture pixels are expressed in a summation of residual errors and High-order moments of the spectral components.

One of their applications is the detection of sub-pixels of an HSI.

## V. FACTORS INFLUENCING THE HSI DATA ANALYSIS

Two main factors influence the analysis of hyperspectral imaging data: the number of training pixels, and the definition of classes.

### A. Number of Pixels Training

One of the problems encountered in data mining analysis (case of HSI) is the absence or low number of training pixels [14]:
- This number has an impact on the number of selectedbands to have better performance.
- For each number there will be a number of bands beyond which the performance degrades.
- To keep good performances, we have to consider aninfinite number of training examples.

Fukunaga [15] shows that the number of training examples is linearly dependent on the dimensionality for linear classifiers; it depends on the square of the dimensionality for the quadratic classifier. Experiments, Lee [16], indicate that second order statistics are more discriminatory for high dimensional data.

### B. *Defining Classes*

The issue of defining classes arises when the ground truth is not given, or incomplete, leading to define other classes. We briefly report that there are three conditions for an optimal class definition [17]:
1) A class must be Exhaustive.
2) A class must be separable.
3) Their values must be Informative.

## VI. APPLICATIONS OF HSI

The hyperspectral remote sensing technology has reached spectacular advances in acquisition of high dimensional data with a higher spectral resolution, thereby increasing the discrimination of spectral signatures compared to traditional multispectral sensors. Thus, it has been used as an important means for Earth observation and exploration, the study of plant's stress which reduces the food performance [18], the exploration of the Moon, Mars and other planets, and also sort mineral and non- mineral waste for recycling [19]. It is an economic technology to provide useful and necessary information on land resources, both for industrial applications and for scientific interests [10] and defense. Hyperspectral imaging has also been crucial to help the food processing industry [20], detection of contaminants in food processing [21].

Data mining is also found among the applications of HSI. This brings us to the investigation of this relationship in the following section.

## VII. RELATIONSHIP BETWEEN HSI AND DATA MINING

Pattern recognition has been initiated from the pioneers works of Frank Rosenblatt on Perceptron [22]. It has been extended for a long time in other directions and has developed into a separate discipline. One of the characteristics of HSI is the large amount of data and the high dimensionality of the vectors manipulated. This high dimensionality of the data provides more capacity of discrimination in classification, but also imposes high-Cost Calculator and the data modeling becomes more complex. This presents a challenge for analysis methods and leads to the use of data mining in many tasks as attributes selection. Fayyad at [23] introduces the data processing chain. Thus, we can see that the HSI must undergo similar patterns to derive knowledge as a thematic map Figure.3 [24]. In what follows, we present the main concepts and definitions of data mining in connection with HSI.

### A. Purpose of Data mining

Consider a set of data vectors with multidimensional numerical attributes. By studying the clusters formed by these vectors, we can discover some hidden behavior in data. Cheng
[24] refers to the need to use specific algoritluns to perform this task on sub sets, high data dimensionality, and it marks the need for dimensionality reduction, taking into the fact thatif A, B and C are disjoint subsets of data sets and a pattern canexist in A and B, but C is independent of A and B, then C is adisturbance against the detection of the concerned pattern. The interpretation of data in high dimensionality is sensitive, and it is preferable to perform pattern recognition in suitable dimensions.

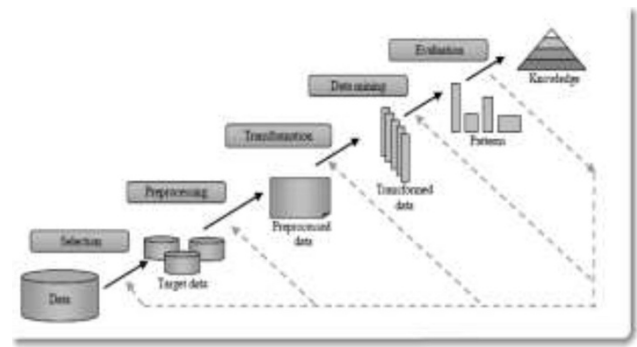

Fig. 3. Data mining Process

In particular, in the context of classification problems, the selection of data can provide an improved accuracy (as compared to standard techniques single-sensor single-date applied to images), which can be of paramount importance in real applications. In hyperspectral imagery, manipulated vectors have dimensions that can be treated in terms of technique of data mining.

In data mining, we can differentiate between the Model and Pattern. A model is a comprehensive description of the data. By against the pattern is a local description that may involve some data attributes [25]. the detection of a certain pattern between two variables does not imply the existence of a causality between them: for example, the existence of a linear model between buyers of a type of drink and buying clothing brand is very interesting for marketing perspective, but it we cannot manipulate a variable by the action on the other. A pattern "A" is a more general than a pattern "B" if whenever B is realized in the data, "A" also occurs. For example, the pattern "At least five types of vegetation exist in the map " is more general than the model "At least two types of vegetation exist in the map." The use of such a generalization between patterns leads to simple algorithms to find all patterns of a certain type that occur in the data. The patterns, such as cores, are often described as non- parametric, because the model is largely driven by data without any parameter in the conventional sense. These smoothing techniques (such as kernel-based models) are useful for the interpretation of data,at least in the case of one or two dimensions. However, no model provides an answer to all problems, and local kernel-based model have weaknesses. In particular, when the number of predictor variables increases, number of data points required to provide accurate estimates increases exponentially (a consequence of the curse of dimensionality). Also, we should note that for large dimensions, we lose the interpretability of models.

B. *Description Models and Predictive Models*

The descriptive models are sufficient to summarize the data in a way to understand the operation of the process. In contrast, predictive models specifically intended to enable to predict the unknown value of a variable of interest, given the other variables. In the context of hyperspectral imagery classification, we are interested in the type of nominative variables of interest to indicate a class devoid of any digital sense. However, there are cases where the predictor variables have numerical meaning such as energy consumption over the next two years.

In a predictive model, one variable is expressed in terms of another, for example we can predict by past customer behavior, the probability of having a new loan. if there is quantitative this Application is a regression, if there is qualitative then the application is a classification, and it's the case of hyperspectral images.

C. *Components of an Algorithm in Data mining*

Given the overall goal in the HSI classification, which is patterns matching that represent the classes to be assigned to test pixels, especially the pixels that are not labeled, it is interesting to indicate that an algorithm in data mining must contain five major components [25]:

1) The Task: visualization, classification, clustering, regression, feature selection and so on.

2) The Model or Structure designs (patterns) determine the structural or functional underlying forms that we seek from the data.

3) The score function: in an ideal world, the choice of the score function accurately reflects the value (the true expected benefit) of a particular predictive model.

4) Research methods and optimization: the goal of optimization and research is to determine parameter values that achieve a minimum (or maximum) of the score function. The search for the "best" values of the parameters is a optimization problems.

5) Data management strategy

From the standpoint of synthesis, by combining various components in different combinations, we can construct data mining algorithms with different properties.

D. *Partial Conclusion*

In this section, we put in a situation the analysis of HSI as an action of data mining. Indeed, as we specify below, the HSI classification requires the resolution of the problem of dimensionality reduction.

VIII. ISSUES RELATING TO HIS

The increased number of spectral bands in the hyperspectral imaging, allows a priori to increase separability between classes, but the accuracy of the statistical estimation decreases with the size of the gap leading to have poor classification results (the curse of dimensionality) [14]. And several approaches to reduce dimensionality of HSI are present in the literature. This size reduction is justified by the properties of high dimensional spaces but also by large duplication of existing information between adjacent spectral bands. To address the fact that certain parts of the spectrum will provide a much richer descriptor for classification than others, some approaches rely on the selection of attributes, others employ the extraction of attributes or combination of selection and data extraction.

Dimensionality reduction of HSI is an important preprocessing for the data analysis of hyperspectral images
[26] from the user side. This pretreatment also includes technical data extraction as selecting attributes and aims to answer the question how to take advantage of reliable and efficient way of using hyperspectral data: The large amount of data involved in hyperspectral imaging significantly increases the time and the complexity of treatment. Effectively reduce the amount of data or select the corresponding bands associated with a particular application of the data set becomes a priority task for the analysis of hyperspectral images [27].

IX. CONCLUSION

The hyperspectral remote sensing is a tremendous leap in the field of remote sensing. The increasing availability of hyperspectral data has enriched us with better data quality and allow us a much stronger ability to identify substances. However, approaches for identifying attributes of hyperspectral images are not as successful as we thought.

Too many bands and a large amount of data not only cause difficulties of data storage and transmission, but also news challenges in technology processing of hyperspectral image: selection and transformation channel; and especially pattern recognition in hyperspectral image attributes. The classification with Hyperspectral images is a rich technology in terms of multidisciplinary applications. But it has charges related to the acquisition and image processing in the higher dimensionality, with the exigencies of data mining technics. In this paper, we are particularly interested in the aspect of data processing. Twoissues related to the processing of hyperspectral data were highlighted: How to eliminate redundancy bands (attributes or features) and select or extract the relevant bands.